%% file: root.tex
\documentclass[letterpaper, 10 pt, journal, twoside]{IEEEtran}

\IEEEoverridecommandlockouts                              

\usepackage{amsmath}
\usepackage{xspace}

\usepackage[utf8]{inputenc} 
\usepackage[T1]{fontenc}    
\usepackage{hyperref}       
\usepackage{url}            
\usepackage{booktabs}       
\usepackage{amsfonts}       
\usepackage{nicefrac}       
\usepackage{microtype}      
\usepackage[dvipsnames]{xcolor}
\usepackage{graphicx}
\usepackage{subfigure}
\usepackage{amsthm}
\usepackage{amssymb}
\usepackage{wrapfig}
\usepackage{dsfont}
\usepackage[ruled, noend]{algorithm2e}
\usepackage{mathrsfs}
\usepackage{multicol}
\usepackage{threeparttable}
\usepackage{multirow}
\usepackage[dvipsnames]{xcolor}
\usepackage[export]{adjustbox}
\usepackage{cite}

\definecolor{myred}{rgb}{0.831, 0.169, 0.090}
\definecolor{mygreen}{rgb}{0.471, 0.710, 0.306}
\definecolor{myblue}{rgb}{0.275, 0.494, 0.831}

\newtheorem{definition}{Definition} 
\newcommand{\revise}[1]{\textcolor{black}{#1}}

\title{\LARGE \bf
Safety-aware Causal Representation for Trustworthy Offline Reinforcement Learning in Autonomous Driving}

\author{
\thanks{Manuscript received: October 29, 2023; Revised February 4, 2024; Accepted February 26, 2024. }
\thanks{This paper was recommended for publication by Editor Jens Kober upon evaluation of the Associate Editor and Reviewers' comments.} 
Haohong Lin$^{1}$, Wenhao Ding$^{1}$, Zuxin Liu$^{1}$, Yaru Niu$^{1}$, Jiacheng Zhu$^{1}$, \\ Yuming Niu$^{2}$ and Ding Zhao$^{1}$ 
\thanks{$^{1}$Haohong Lin, Wenhao Ding, Zuxin Liu, Yaru Niu and Ding Zhao are all with the Department of Mechanical Engineering, Carnegie Mellon University, Pittsburgh, PA 15213 USA. 
{\tt\small \{haohongl, wenhaod, zuxinl, yarun, jzhu4\}@andrew.cmu.edu, dingzhao@cmu.edu},}%
\thanks{$^{2}$Yuming Niu is with the Ford Motor Company, Dearborn, MI 48126 USA.  
{\tt\small \{yniu4\}@ford.com},}%
\thanks{Digital Object Identifier (DOI): see top of this page.}
}

\begin{document}

\maketitle

\begin{abstract}

\revise{In the domain of autonomous driving, the offline Reinforcement Learning~(RL) approaches exhibit notable efficacy in addressing sequential decision-making problems from offline datasets. However, maintaining safety in diverse safety-critical scenarios remains a significant challenge due to long-tailed and unforeseen scenarios absent from offline datasets. In this paper, we introduce the saFety-aware strUctured Scenario representatION (FUSION), a pioneering representation learning method in offline RL to facilitate the learning of a generalizable end-to-end driving policy by leveraging structured scenario information.} FUSION capitalizes on the causal relationships between the decomposed reward, cost, state, and action space, constructing a framework for structured sequential reasoning in dynamic traffic environments. We conduct extensive evaluations in two typical real-world settings of the distribution shift in autonomous vehicles, demonstrating the good balance between safety cost and utility reward compared to the current state-of-the-art \revise{safe RL and IL baselines}. Empirical evidence in various driving scenarios attests that FUSION significantly enhances the safety and generalizability of autonomous driving agents, even in the face of challenging and unseen environments. Furthermore, our ablation studies reveal noticeable improvements in the integration of causal representation into \revise{the offline safe RL algorithm.
Our code implementation is available on the \href{https://sites.google.com/view/safe-fusion/}{project website}.
}

\end{abstract}

\begin{IEEEkeywords}
Intelligent Transportation Systems, Representation Learning, Reinforcement Learning
\end{IEEEkeywords}

\input{1_introduction.tex}
\input{2_preliminary.tex}
\input{3_methodology.tex}

\input{4_experiments.tex}

\input{5_conclusions.tex}

\section*{ACKNOWLEDGEMENT}
The authors gratefully acknowledge the support from the National Science Foundation under grants CNS-2047454 and gift funding from Ford Motor Company.
\bibliographystyle{IEEEtran}
\bibliography{reference}



\end{document}

%% file: 1_introduction.tex
\section{Introduction}
\label{sec:intro}
\IEEEPARstart{L}{earning} \revise{from Demonstration~(LfD) techniques have achieved huge success in autonomous driving~\cite{chen2019model, pan2020imitation, chen2021interpretable} by improving the representation quality in an end-to-end framework. Among all the solutions categorized as LfD, Offline Reinforcement Learning has shown its superiority in many other robotic tasks, including locomotion and manipulation~\cite{rafailov2021offline}. 
However, in the context of autonomous driving, the safety and generalizability of learning-based policies in various safety-critical scenarios remain elusive~\cite{ding2023causalaf, renz2022plant, shao2023safety}. The distribution shift between offline training samples and online testing environments makes it harder to deploy the learning algorithms to the online environments safely. }
Prior studies~\cite{ding2023survey, xu2022trustworthy} illustrate that even minor domain shifts in road structures or surrounding vehicles can result in catastrophic outcomes due to the high-stakes nature of autonomous driving.

\begin{figure}[t]
  \centering
  \includegraphics[width=0.50\textwidth,left]{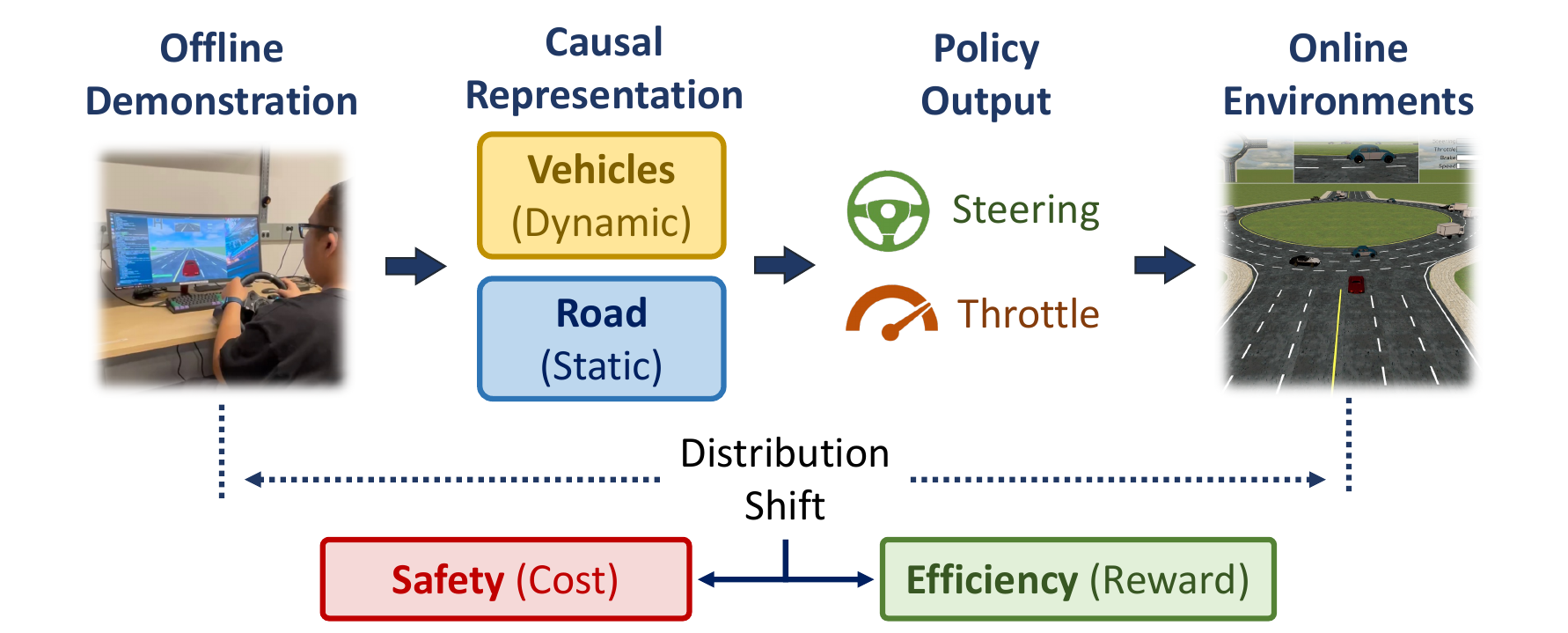}
  \vspace{-5mm}
  \caption{Diagram depicting offline-to-online generalization via a modular reasoning framework. The agent learns a causal abstraction from offline demonstration trajectories and then applies it to different environmental components during online implementation. \revise{The distribution shift between offline datasets and online environment can lead to unsatisfying safety or efficiency in driving performance.} This abstracted representation enables learning agile agents for unseen scenarios in a zero-shot manner while enhancing safety and efficiency.}
  \label{fig:motivation}
  \vspace{-5mm}
\end{figure}

Although existing research has successfully applied end-to-end learning-based algorithms to racing cars~\cite{fuchs2021super, wurman2022outracing, shah2023fastrlap}, urban driving scenarios remain complex for existing learning-based agents. Complexity arises from the fact that urban settings require structural reasoning in context-rich and safety-critical situations~\cite{mohan2023structure}. For instance, \textit{humans} can effortlessly adapt their driving behaviors based on static contexts such as roadblocks or dynamic contexts such as surrounding traffic, often making intuitive judgments, as illustrated in Figure.~\ref{fig:motivation}. Although such \revise{causal abstraction} is straightforward to humans with high reasoning capabilities, end-to-end approaches, such as vanilla deep RL methods, usually fail due to the distribution shift in various driving scenarios \revise{neglecting the underlying structures of the scenarios and usually resulting in being over-conservative or over-aggressive}. As a consequence, two pivotal challenges emerge under such distribution shifts: (i) ensuring safety performance unseen driving contexts and (ii) striking a balance between safety and driving efficiency.

\revise{Recent LfD advances in autonomous driving improve the trustworthiness of learned policies through representation learning in offline RL or IL, including the object-centric representation~\cite{renz2022plant}, safety-enhanced scene representation~\cite{shao2023safety, jia2023think}, multi-modal sensory representation~\cite{lee2019ensemble}, domain-invariant state representation~\cite{bica2021invariant}, agile action abstraciton~\cite{pan2020imitation}, and hierarchical action representation~\cite{akrour2018regularizing}.} \revise{However, a recurring limitation of these representation learning works is the assumption of access to perfect expert demonstrations, which may not be accessible in diverse urban scenarios.}

To mitigate the reliance on perfect expert demonstrations, multiple offline RL~\cite{kumar2019stabilizing, fujimoto2019off} and safe RL~\cite{xu2022constraints, liu2023constrained} approaches have been proposed. 
These methodologies harbor the potential to equilibrate the RL agents' priorities between safety and efficiency, especially when learning from non-expert demonstrations. Encouragingly, some studies~\cite{fang2022offline, shah2023fastrlap} manage to surpass expert policies during online deployment by using these batch RL methods, which are based on improved real-world data. 
Although these works overcome the limitation of perfect expert demonstration, they mostly assume that online environments will mirror the dynamics of those from which offline trajectories were collected. 
\revise{In reality, the scarcity and lack of diversity of high-quality expert data always exist and lead to significant distribution mismatch between training and deployment. This is particularly apparent in autonomous driving, where static (e.g., road layouts) and dynamic (e.g., traffic flow) contexts differ markedly across locales. How to achieve generalizability in unseen scenarios remains an open research question. }

In this study, we introduce sa\textbf{F}ety-aware str\textbf{U}ctural \textbf{S}cenario representat\textbf{ION} (FUSION), which aims to improve the generalizability of the safety performance of self-driving cars \revise{under distribution shift}. More concretely, our contributions are summarized as follows: 
\begin{itemize}
\item \revise{We introduce a safety-aware offline reinforcement learning framework that aims to improve generalizability under distribution shifts during the online deployment stage. 
\item We develop a self-supervised causal representation learning paradigm to regularize the scenario representation, encouraging a better balance between the safety and efficiency of the learned policies. }
\item We provide comprehensive evaluations on the offline dataset collected from the human beings and Intelligent Driver's Model~(IDM), \revise{showing the advantage of FUSION over the existing state-of-the-art approaches in offline safe RL~\cite{liu2023datasets} and IL-based methods~\cite{lee2019ensemble, bica2021invariant, akrour2018regularizing}.}
\end{itemize}

%% file: 2_preliminary.tex
\section{Related Works}

\textbf{Safety-aware Decision Making from Offline Data.}
To bring up safety awareness of autonomous vehicles, the most recent works formulate the safe decision-making problem as constrained optimization~\cite{achiam2017constrained,liu2022constrained, shao2023safety}. 
However, there have been several different roadmaps for solving this problem. 
For the IL-based approach, ~\cite{menda2019ensembledagger, lee2019ensemble} propose implicit safe constraints in IL via uncertainty quantification and Bayesian abstraction from expert data. These approaches depend on their \textit{ safety} on the small discrepancy between the learned trajectory and the expert trajectory. More explicitly, InterFuser~\cite{shao2023safety} proposes a safe controller that utilizes interpretable intermediate features to directly constrain the controller output within a safety set. 
On the other hand, offline Reinforcement Learning (RL) agents manage to balance safety and efficiency with additional information on the reward, cost, and cost threshold along the trajectories~\cite{le2019batch, liu2023datasets}. 
To fully extract temporal information from offline trajectories, recent works turn offline RL into a sequential modeling problem using the power of transformers~\cite{chen2021decision, liu2022augmenting, liu2023constrained, janner2021offline}. 
Most of these works ignore the inherent structures of MDP in either the spatial or temporal domain, which limits the generalizability of the policy.

\textbf{State Abstraction for Decision Making.}
To improve the performance of decision-making agents with some extra structural information, some recent work has focused on deriving state abstraction for generalizable decision-making using representation learning tricks. 
In the IL realm, ~\cite{bica2021invariant} proposes Invariant Causal Imitation Learning~(ICIL) to deal with the distribution shift with domain-invariant causal features. 
Based on uncertainty quantification, \cite{lee2019ensemble, menda2019ensembledagger, loquercio2020general} propose an ensemble representation that leverages multi-modal sensor inputs to improve generalizability for self-driving agents. 
PlanT~\cite{renz2022plant} proposes a learnable planner module based on object-centric representations. 
The RL field has seen developments in state abstraction through self-supervised learning methods, including time-contrastive learning~\cite{sermanet2018time}, hierarchical skill decomposition~\cite{akrour2018regularizing} and deep bisimulation metric learning~\cite{zhang2020learning, dadashi2021offline}. 
In autonomous driving applications, state and action space are usually factorizable, ~\cite{ding2022generalizing, ding2023seeing} propose to train RL agents under the guidance of causal graphs to improve generalizability by discovering the latent structure in the world or policy model. 
\revise{Prior to this work, however, the intersection of state abstraction with offline Safe RL is unexplored, which is crucial to advance the learning-based methods in the autonomous driving domain.}

\begin{figure*}[h]
  \includegraphics[width=0.98\textwidth]{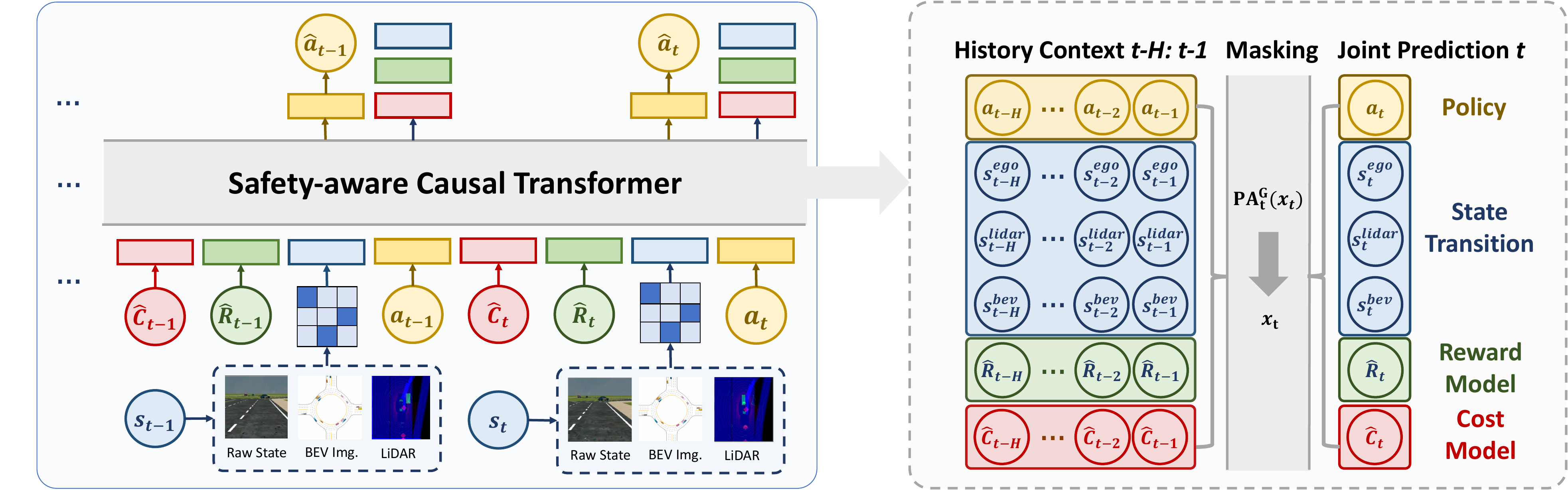}
  \caption{Overview of Safety-aware structural Scenario Representation Framework. The diagram on the left shows a safety-aware decision transformer that conducts sequential decision-making based on the temporal contexts. The right diagram shows the general form of the graphical model in the CEWM and Policy Learning modules in FUSION, where the connection between different timesteps will be determined by the attention weights in the causal transformer. The nodes in a later timestep depend on their parental nodes in the previous timesteps.} 
  \label{fig:fuse}
\end{figure*}

\section{Problem Formulation}

\revise{As stated in Section~\ref{sec:intro}, this work essentially aims to tackle a generalizable safe RL problem under distribution shifts in an offline setting.} To better model such distribution shifts, we follow the definition of contextual MDP in~\cite{chen2021context} to define the Constrained Contextual Markov Decision Process, or $C^2$-MDP, to model this generalizable safe RL problem as follows: 

\begin{definition}
\label{def:C2-MDP}
Constrained Contextual Markov Decision Process~($C^2$-MDP) is a Contextual MDP with a tuple \revise{$\big(\Omega, \mathcal{M}(\omega)\big)$}, where $\mathcal{M}$ is a function that maps any contexts $\Omega \in \Omega$ to a constrained MDP 
\revise{$\mathcal{M}(\omega) = \big( \mathcal{S}, \mathcal{A}, P_\omega, r, c, s_0, \gamma \big)$. }
\end{definition}

Here \revise{$P_\omega: \mathcal{S}\times \mathcal{A}\times \Omega \to \mathcal{S}$} is the context-specific transition function, $r: \mathcal{S}\times \mathcal{A}\to \mathbb{R}$ is the reward function, $c: \mathcal{S}\times \mathcal{A}\to \mathbb{R}$ is the cost function, $s_0$ is the initial state, and $\gamma$ is the discount factor. 
$C^2$-MDP defines the safety cost as an additional intuitive performance preference for driving agents. In addition, it includes different MDPs according to different contexts $\omega$. This additional context aims to model the phenomena that the traffic environment varies across different contexts (e.g. road types or traffic densities) in the autonomous driving scenarios. 

Following the above definition, we introduce our problem formulation and then give a sketch of our proposed learning pipeline for generalizable safe RL problems in autonomous driving. 
Based on the Definition~\ref{def:C2-MDP}, the Constrained Contextual MDP aims to maximize cumulative reward while satisfying the safety constraints on cumulative expected cost under a certain target context $\omega$. In formal terms, our problem can be defined as the following constrained optimization problem $\max_\pi  \ J_r(\pi, \omega) \quad s.t.  \ J_c(\pi, \omega)\leq \kappa_c$, where we define the reward objective
$J_r(\pi,\omega)\triangleq \mathbb{E}_{\omega, \pi} \sum_{t=1}^T r(s_t, a_t)$
and similarly the cost objective, $J_c(\pi, \omega)=\mathbb{E}_{\omega, \pi}\sum_{t=1}^T c(s_t, a_t)$. 

To achieve generalizable safety, we aim to optimize a policy that satisfies safety constraints: $J_c(\pi, \omega) \leq c, \forall \pi\in \Pi, \omega\in \Omega$, 
i.e. imposing constraint satisfaction under varying behavior policies $\pi_\beta$ and environment contexts $\omega$. Meanwhile, we assume that the preference for the reward function $r$ and the cost function $c$ remain unchanged in different contexts. 

In our autonomous driving problem, the reward is composed of a forwarding reward in the longitude direction, a continuous reward for the vehicle speed, and an additional sparse reward once the vehicles reach the goal or other terminal states: 
\begin{equation}
\label{eq:reward}
\begin{aligned}
    r_t & = w_1^r r_{\text{forward}}+w_2^r r_{\text{speed}}+w_3^r r_{\text{term}} \\
     & = w_1^r (d_t-d_{t-1})+w_2^r v_t+w_3^r \mathbb{I}(s_t=g) 
\end{aligned}
\end{equation}
In our urban driving task, \revise{the safety cost comes from three events: (i) collision with others, (ii) out-of-road, and (iii) over-speeding}. Collision and out-of-road costs are binary indicators \revise{that are non-zero only when the corresponding event happens}, and overspeeding costs are a continuous cost that occurs once the vehicle exceeds a certain speed limit $v_{\text{limit}}$. 
\begin{equation}
\label{eq:cost}
\begin{aligned}
    c_t &=w_1^c c_{\text{collision}}+w_2^c c_{\text{out road}}+w_3^c c_{\text{overspeed}} \\
    = & w_1^c \mathbb{I}(s\in s_{\text{collision}}) +w_2^c \mathbb{I}(s\notin s_{\text{road}})+w_3^c \max(0, v_t-v_{\text{limit}})
\end{aligned}
\end{equation}

\revise{The core problem formulation in this paper is to learn a safe policy with good generalizability at the deployment stage, under distribution shifts that occur: }
(i) between offline data collected from mixed-quality policies and online environments, i.e. $\pi_\beta\neq \pi^*$, and 
(ii) between varying contexts of $C^2$-MDP, i.e. training environments $\omega_{train}$ for data collection are different from online testing environments $\omega_{test}$. 
This difference also indicates the difference in MDP $\mathcal{M}(\omega_1)\neq \mathcal{M}(\omega_2)$. More specifically, we define the distribution shift in \textit{transition dynamics} $T_\omega$ (e.g., the density of traffic) as follows: $p(\cdot|s,a; \omega_{train})\neq p(\cdot|s,a; \omega_{test})$. 

%% file: 3_methodology.tex
\section{Methodology}

In this section, we zoom in on more details about our proposed FUSION with two important modules: (i) Causal Ensemble World Model~(CEWM), and (ii) safety-aware Causal Bisimulation Learning~(CBL).

\subsection{Causal Ensemble World Model Learning} 
In autonomous driving problems, the entire state space can be decomposed into several disjoint subspaces~\cite{lee2019ensemble}, including the estimated ego navigation state, lidar observation, and visual observation, e.g. the birds-eye-view observation that serve as input to FUSION in Figure~\ref{fig:fuse}.

\begin{definition}[Factorizable State Space]
    The factorizable state space in MDP indicates a disjoint state space decomposition, where $S=S_1\cup S_2\cup \cdots \cup S_N$, and $N$ indicates how many disjoint state components we have in a certain problem. 
    \label{def:factorize}
\end{definition}

To help the FUSION framework gain better awareness of the structure of the state and action space, we propose the CEWM based on multi-modal observations, as defined 
The factorized state space Definition~\ref{def:factorize}, along with the reward, cost, and action variables, form the nodes in this world model. To better describe the structural dependency between them, we further design the CEWM according to the following definition of Structured Causal Model~(SCM).

\begin{definition}
An SCM $(\mathcal{S}, \mathcal{E})$ consists of a set of variables $\mathcal{S}$, along with $d$ functions~\cite{peters2017elements},
\begin{equation*}
    s_j := f_j(\mathbf{PA}^{\mathcal{G}}(s_j), \epsilon_j),\ \  j \in [d],
\end{equation*}
where $\mathbf{PA}_j^\mathcal{G} \subset \{ s_1,\dots,s_d \} \backslash \{s_j \}$ are called parents of $s_j$ in the Directed Acyclic Graph~(DAG) $\mathcal{G}$, and $\mathcal{E} = \{ \epsilon_1,\dots,\epsilon_d \}$ follows a joint distribution over the noise variables, which are required to be jointly independent. 
\label{def_scm}
\end{definition}

For general offline RL problems, SCM aims to jointly parameterize the world model and the policy model between different nodes in the state, action, reward, and safety cost. To parameterize the functions $f$ in this SCM, we use a Safety-aware Causal Transformer, as shown in Figure~\ref{fig:fuse}. 
For example, the child node $s_j$ is determined by its parent tokens $\mathbf{PA}_t^{\mathcal{G}}(s_j)$ in the previous tokens $\tau_{t-H:t}$, and the exogenous noise variable $\epsilon_j$, which are aggregated by a variable-specific function $f_j$ empowered by the attention mechanism of Transformer. 
The edges between different nodes represent their causal dependency in the spatio-temporal domain, which is essentially captured by the attention weights, as we will discuss later in Figure~\ref{fig:attn} of the experiment parts. 
In addition to capturing the cause-and-effect relationship between the reward, cost, and factorizable state space, the SCM also enjoys a great property in that the child nodes (e.g., the state and reward/cost in subsequent timesteps) are only dependent on their parent nodes~(in the state or action space in the previous timesteps) while removing unnecessary dependencies between the descendent nodes to indirect ancestors or non-parent nodes. 
This property improves both generalizability and efficiency for an autoregressive inference during the online deployment.

Based on this property, we derive the CEWM under the SCM, which can then be decomposed into the following disjoint components, including the reward-to-go model, cost-to-go model, the factorized state-action transition dynamics, and the policy optimization, as is shown below:  
\begin{equation}
    \begin{aligned}
    p (\tau_{t}| &\tau_{t-H:t}) = p(a_{t}, s_{t}, R_{t}, C_{t} | a_{t-1}, s_{t-1} \cdots R_{t-H}, C_{t-H}) \\ 
      = & \underbrace{p\Big(r_t| \mathbf{PA}_t^G(r_t)\Big)}_\text{Reward-to-go}  \underbrace{p\Big(c_t| \mathbf{PA}_t^G(c_t)\Big)}_\text{Cost-to-go} \\
      & \underbrace{ p \Big(a_{t+1} | \mathbf{PA}_t^G(a_{t+1})\Big)}_\text{Policy Optimization}  \underbrace{\prod_{i\in \dim(S)} p \Big(s_{t+1}^i | \mathbf{PA}_t^G(s_{t+1}^i)\Big)}_\text{Factorized Dynamics}
    \end{aligned}
\label{eq:decomposition}
\end{equation}

Therefore, we exert an auxiliary task of trajectory optimization in the optimization process of safety-aware decision transformer to estimate the three components in (\ref{eq:decomposition}), i.e. 
\begin{equation}
    \begin{aligned}        
    \mathcal{L}_{\text{traj}} & = -\log p(\tau_{t+1}|\tau_{t-H:t}) =  -\log p(R_t| \mathbf{PA}_t^G(R_t))  \\
    & - \log p(C_t| \mathbf{PA}_t^G(C_t)) - \log p (a_{t+1} | \mathbf{PA}_t^G(a_{t+1}))  \\
    & - \sum_{i\in \dim(S)}\log p (s_{t+1}^i | \mathbf{PA}_t^G(s_{t+1}^i)) \\ & =\underbrace{\mathcal{L}_{\text{rtg}}}_\text{Reward Critic} + \underbrace{\mathcal{L}_{\text{ctg}}}_\text{Cost Critic} + \underbrace{\mathcal{L}_{\text{act}}}_\text{Policy Optimization} +\underbrace{\mathcal{L}_{\text{dyn}}}_\text{Transition Dynamics}
    \end{aligned}
\label{eq:trajectory}
\end{equation}

This trajectory optimization objective benefits our safety-aware DT with better structural awareness of the trajectory level between state, action, reward-to-go, and cost-to-go. 
The design of this safety-aware DT model manages to parameterize the CEWM that we propose in~(\ref{eq:decomposition}), as the latter token is generated conditioned on the previous tokens in an auto-regressive way. 

\begin{figure}[t]
  \centering
  \includegraphics[width=0.5\textwidth]{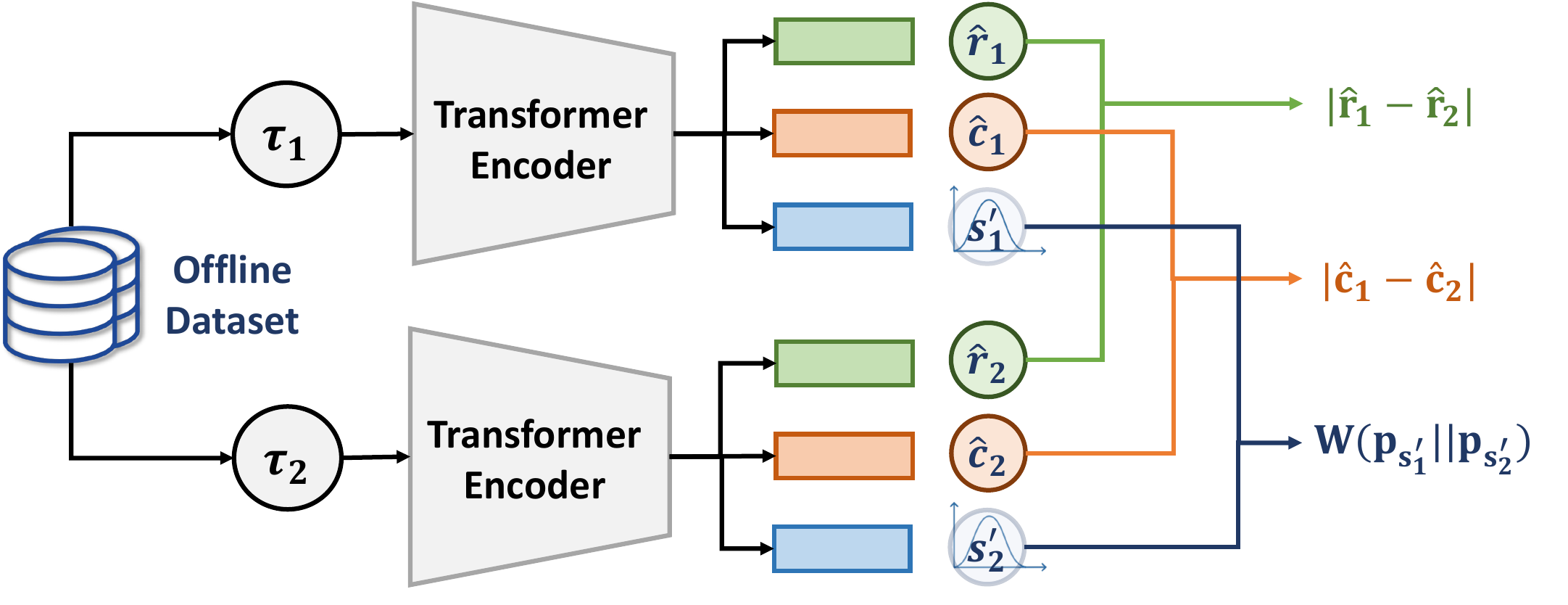}
  \vspace{-6mm}
  \caption{Safety-aware bisimulation metrics with the distribution distance in transition dynamics, rewards, and safety cost.}
  \label{fig:bisim}  
  \vspace{-5mm}
\end{figure}


\subsection{Safety-aware Bisimulation Learning}
Though CEWM provides an \textit{explicit} structure to model the causality, learning such a model from offline datasets is non-trivial. 
The reason is that demonstrations in the mixed-quality dataset have diverse levels of safety due to spurious correlations between actions and states.
To avoid getting misled by such spurious correlation, we introduce an additional self-supervised regularization term in an \textit{implicit} way, namely Causal Bisimulation Learning, or CBL. 
Inspired by the DBC algorithm for off-policy RL in~\cite{zhang2020learning}, we further regularize the FUSION model with safety-aware Bisimulation Learning in our offline RL setting. 
We first extend the traditional bisimulation relationships for MDP in~\cite{li2006towards, zhang2020learning} with an extra safety term: 

\begin{definition}[Safety-aware Bisimulation Relation]
A safety-aware bisimulation relation $\mathcal{U}\subset \mathcal{S}\times \mathcal{S}$ is a binary relation which satisfies,
$\forall (s_1, s_2)\in \mathcal{U}$: 
\begin{itemize}
    \item $\forall a\in \mathcal{A}, r(s_1,a)=r(s_2,a)$
    \item $\forall a\in \mathcal{A}, c(s_1,a)=c(s_2,a)$
    \item $\forall a\in \mathcal{A}, s'\in \mathcal{S}, p(s'|s_1,a) = p(s'|s_2,a)$. 
\end{itemize}
\end{definition}

Intuitively, in the Constrained MDP setting, the bisimilarity between two states is determined not only by the stepwise reward and transition dynamics but also by their similarity in the step-wise cost. 
In practice, the reward, cost, and transition dynamics could hardly match exactly for two different states, therefore, we propose a smooth alternative~\cite{ferns2004metrics} of safety-aware bisimulation relationship, denoted as Safety-aware Bisimulation Metrics as is shown in Figure~\ref{fig:bisim}. 
\begin{definition}[Safety-aware Bisimulation Metrics]
The bisimulation metric $d^\pi: \mathcal{S}\times \mathcal{S}\to \mathbb{R}^+$ is a mapping from the state space to a non-negative scalar, defined as:  
\begin{equation} 
\begin{aligned}
     & d^\pi(s_1, s_2) =  \mathbb{E}_{a_1\sim \pi(\cdot|s_1),\atop a_2\sim \pi(\cdot|s_2)} \Big[|r(s_1, a_1)-r(s_2, a_2)|  \\
     & + \lambda |c(s_1, a_1)-c(s_2,a_2)|  + \gamma W_2(\hat{p}(\cdot|s_1,a_1), \hat{p}(\cdot|s_2,a_2))\Big], \\
\end{aligned}
\label{eq:bisim}
\end{equation}
\end{definition}
The Lagrangian multiplier $\lambda$ aims to balance the safety term, and $W_2(\cdot,\cdot)$ is the 2-Wasserstein distance measuring the similarity between two transition dynamics distributions. 
We use the following learning objectives to align the state representation with the bisimulation metrics in the latent space: 
\begin{equation}
\label{eq:bisim_loss}
    \begin{aligned}
    \mathcal{L}_{\text{bisim}}=\mathbb{E}_{s_1, s_2\sim p_{\pi_\beta}} \Big(\|\phi(s_1)-\phi_{sg}(s_2)\|_1 - d^\pi(s_1, s_2)\Big)^2,
    \end{aligned}
\end{equation}
where $\phi_{sg}$ means stop gradient of state encoder $\phi$.



Finally, at inference time, we take advantage of the prediction of values in online inference time, as shown in Algorithm~\ref{alg:fusion_inference}. By taking the minimum cost-to-go preference and cost prediction, and the maximum reward-to-go preference and reward prediction at each step, we aim to improve the safety and efficiency of FUSION conditioned on the input human preference during the online deployment stage.

\newcommand\mycommfont[1]{\small\ttfamily{#1}}
\SetCommentSty{mycommfont}
\begin{algorithm}[htp]
\caption{Safety-aware CBL}
\label{alg:bisim}
\KwData{Offline (mixed) trajectories, cost limit $C$}
\KwResult{State encoder $\phi$ of policy $\pi$}

\For{$k=0,\cdots, N-1$} {            
    Sample minibatch: $\mathcal{B}\gets \text{Sample}(\mathcal{D}_{\pi_\beta})$\;
    Construct transition pairs: $(s_1, a_1, s_1')\gets \mathcal{B}$ \;
    \textbf{Permute samples}: $(s_2, a_2, s_2')\gets \text{permute}(\mathcal{B})$ \;
    \textbf{Compute bisimulation distance: } With (\ref{eq:bisim}) \;
    \textbf{Update encoder: } $\phi_{k+1} \gets \phi_k - \nabla_\phi \mathcal{L}_{\text{bisim}}$ with~(\ref{eq:bisim_loss})\;
}
\end{algorithm}
\begin{algorithm}[htp]
\caption{Training and Inference of FUSION}
\label{alg:fusion_inference}
\KwData{Context length~$H$, Reward target~$R_0$, Cost limit~$C_0$}
\KwResult{Policy $\pi_{\theta,\phi}$}

\tcc{Offline Training}
\For{$k=0, \cdots, N-1$} {
    \textbf{Update Transformer $\theta$} with CEWM by~(\ref{eq:trajectory})\;
    \textbf{Update Encoder $\phi$} with CBL by Alg.~\ref{alg:bisim}\;
}

\tcc{Online Inference with context $H$} 
$s_0\gets \text{env.reset()}$\; 
$\textbf{o}\gets \{C_0, R_0, s_0\}$\;
$a_0\gets \pi_{\theta,\phi}(\textbf{o})$\;

\For{$t=1, \cdots, T-1$} {
    \text{Rollout:} $s_{t}, r_{t}, c_{t}=\text{env.step}(a_{t-1})$\;
    \text{Predict reward value: } $\hat{R}(s_t, a_t)\gets  \phi^r(s_{t})$\;
    \text{Predict cost value:} $\hat{C}(a_{t}, s_{t})\gets \phi^c(s_{t})$\;
    \text{\small{Update rtg token:}} $R_t\gets \max\{\hat{R}(s_t, a_t), R_{t-1}-r_t\}$\;
    \text{\small{Update ctg token:}} $C_t\gets \min\{\hat{C}(s_t, a_t), C_{t-1}-c_t\}$\;
    \text{\small{Update context: }} $\textbf{o}\gets \{\{a_{t-1}, C_t, R_t, s_t\}\}_{t-H:t}$\;
    \textbf{Predict action: } $a_t\gets \pi_{\theta,\phi}(\textbf{o})$ \;
}
\end{algorithm}

%% file: 4_experiments.tex
\begin{table*}[hbt!]
    \caption{\revise{Evaluation Performance in both policy mismatch and dynamics mismatch settings. Each of the baseline results is evaluated under 5 random seeds. \textbf{Bold} means the best.}}
    \label{tab:main_results}
    \centering
    \begin{tabular}{c | c | c c c c | c c c | c} \toprule
        \textbf{Mismatch} & \textbf{Metrics} & \textbf{Safe BC} & \textbf{ICIL} & \textbf{BNN} & \textbf{GSA} & \textbf{BEAR-Lag} & \textbf{BCQ-Lag} & \textbf{CPQ} & \textbf{FUSION} \\ \midrule
        \multirow{3}{*}{Policy} & Reward~($\uparrow$) & 106.28\scriptsize{$\pm$7.49} & 122.66\scriptsize{$\pm$4.85} & 118.61\scriptsize{$\pm$3.09} & 89.94\scriptsize{$\pm$6.84} & 109.62\scriptsize{$\pm$3.91} & 111.36\scriptsize{$\pm$5.26} & 9.01\scriptsize{$\pm$0.87} & \textbf{139.95\scriptsize{$\pm$4.24}} \\
        & Cost~($\downarrow$) & 12.79\scriptsize{$\pm$0.70} & 11.07\scriptsize{$\pm$1.11} & 4.46\scriptsize{$\pm$0.41} & 13.18\scriptsize{$\pm$1.26} & 4.46\scriptsize{$\pm$0.29} & 0.89\scriptsize{$\pm$0.08} & 1.05\scriptsize{$\pm$0.18} & \textbf{0.52\scriptsize{$\pm$0.06}} \\
        & Succ. Rate~($\uparrow$) & 0.47\scriptsize{$\pm$0.10} & 0.76\scriptsize{$\pm$0.05} & 0.74\scriptsize{$\pm$0.11} & 0.34\scriptsize{$\pm$0.08} & 0.72\scriptsize{$\pm$0.06} & 0.79\scriptsize{$\pm$0.08} & 0.00\scriptsize{$\pm$0.00} & \textbf{0.90\scriptsize{$\pm$0.03}} \\ \midrule
        \multirow{3}{*}{Dynamics} & Reward~($\uparrow$) & 81.07\scriptsize{$\pm$3.80} & 88.21\scriptsize{$\pm$5.30} & 113.35\scriptsize{$\pm$5.68} & 102.40\scriptsize{$\pm$6.44} & 113.38\scriptsize{$\pm$5.25} & \textbf{122.72\scriptsize{$\pm$7.64}} & 7.47\scriptsize{$\pm$0.59} & {117.40\scriptsize{$\pm$4.30}} \\
        & Cost~($\downarrow$) & 9.44\scriptsize{$\pm$0.55} & 7.29\scriptsize{$\pm$0.72} & 19.16\scriptsize{$\pm$0.55} & 11.88\scriptsize{$\pm$0.98} & 7.86\scriptsize{$\pm$0.66} & 6.22\scriptsize{$\pm$0.76} & \textbf{0.71\scriptsize{$\pm$0.09}} & \textbf{0.90\scriptsize{$\pm$0.14}} \\
        & Succ. Rate~($\uparrow$) & 0.12\scriptsize{$\pm$0.06} & 0.32\scriptsize{$\pm$0.05} & 0.59\scriptsize{$\pm$0.06} & 0.03\scriptsize{$\pm$0.02} & 0.32\scriptsize{$\pm$0.05} & 0.39\scriptsize{$\pm$0.08} & 0.00\scriptsize{$\pm$0.00} & \textbf{0.82\scriptsize{$\pm$0.04}}\\
    \bottomrule
    \end{tabular} 
    \vspace{-5mm}
\end{table*}

\section{Experiments}

In this section, we first go through the environments and evaluation protocols that we use based on the MetaDrive simulator~\cite{li2022metadrive}.
Next, we conduct experiments and ablation studies to answer four research questions, aiming to demonstrate how well our proposed methods could learn a safe and generalizable policy based on the offline driver's data. 
The evaluation results illustrate the effectiveness of the FUSION model.

\subsection{Experiment Setup}
\label{sec:env_setup}
\paragraph{Evalation Environment}
We evaluate our algorithm on MetaDrive~\cite{li2022metadrive}, a light-weighted, realistic, and diverse autonomous driving simulator, which can specifically test the generalizability of learned agents on unseen driving environments with its capability to generate an unlimited number of scenes with various road networks and traffic flows. 

The observation of the agents consists of the following components: (i) the ego states and navigation information, which contains the estimation of the ego vehicle's relative position with respect to the closest waypoint for navigation; (ii) the LiDAR observation with 240 laser bins; (iii) \revise{the Birds-eye-view~(BEV) observation, which is an 84$\times$84$\times$5 multi-channel image that captures the road contexts and the recent trajectories of the ego and surrounding vehicles}. 

We collect the offline dataset by IDM polices~\cite{kesting2007general} with diverse levels and styles of aggressiveness of the ego and surrounding drivers. We manually set different acceleration and deceleration rates to adjust the aggressiveness level in the IDM policy. 
\revise{The offline dataset consists of 2,000 trajectories with over 400,000 timesteps under 6 different road configurations. }

\revise{We evaluate the following quantitative metrics to demonstrate the effectiveness of FUSION: }
\begin{itemize}
    \item The \textbf{Utility Reward} metric evaluates the efficacy and efficiency of autonomous vehicles to finish the task, which is a weighted combination of the cumulative driving distance, driving speed, and waypoint arrival, as is introduced in~(\ref{eq:reward}). 
    \item The \textbf{Safety Cost} metric evaluates the overall safety level of autonomous vehicles, which comes from three safety-critical scenarios in autonomous driving, including collision, out-of-lane, and over-speed, as is defined in~(\ref{eq:cost}). The speed limit $v_\text{limit}$ is set to be 40 \textit{kph}. 
    \item The \textbf{Success Rate} metric indicates the ratio of episodes in which the agent successfully reaches the destination within a maximum number of timesteps. 
\end{itemize}

We test our methods in six different types of road configurations (see Figure ~\ref{fig:radar}). 
As introduced in~(\ref{eq:cost}), the safety violation costs are due to three sources: (i) collision, (ii) out-of-lane, and (iii) over-speed. The cost for collision and out-of-lane is 1 at each occurrence, and the over-speed cost $c_{\text{speed}}=\max\{0, 0.02(v-v_\text{limit})\}$. 
An episode will end if any one of the risk scenarios (i) (ii) happens, or the overall timestep is greater than a preset decision horizon of 1,000. 
When the agent reaches the destination without any collision or getting off the road, it will be counted as a success. 

We compare our proposed methods and baselines in the following two settings: 
\begin{itemize}
    \item \textbf{Policy Mismatch} stands for the case where the offline dataset is sampled from the non-perfect expert policy, and the agents need to tackle the generalization challenge from mixed-quality and potentially unsafe offline data towards the deployment in the online environment. 
    \item \textbf{Dynamics mismatch} stands for the case where the agent needs to tackle another generalization challenge from the training environments~(where the offline data is collected) with sparser traffic flows, towards the testing environments where the traffic flows are 1.5$\times$ denser than the training. 
\end{itemize}

\paragraph{Baselines}
We illustrate our results by comparing FUSION against two types of baselines: (i) safe imitation learning and (ii) offline safe reinforcement learning. 
Specifically, the implementation of these baselines aims to solve the multi-modal sensory inputs in the sequential decision-making problems of autonomous driving. 

IL-based methods select safe trajectories or conduct uncertainty quantification to avoid entering uncertain and unsafe regions. This kind of baseline includes Safe Behavior Cloning (\textbf{Safe BC}~\cite{pan2020imitation}) that only uses safe trajectories to train the agent, Invariant Causal Imitation Learning~(\textbf{ICIL}~\cite{bica2021invariant}) that derives invariant state abstraction to learn generalizable policies by the model ensemble, like \textbf{GSA}~\cite{akrour2018regularizing} and \textbf{BNN}~\cite{lee2019ensemble}, which both use hierarchical state abstraction in generalizable decision making. 

On the other hand, offline Safe RL baselines generally solve a constrained optimization problem of $C^2$-MDP by adding Lagrangian terms in the policy evaluation step. Two of them are BEAR Lagrangian~(\textbf{BEAR-Lag}) and BCQ Lagrangian~(\textbf{BCQ-Lag}), which are safety-aware variants of Offline RL algorithms BEAR~\cite{fujimoto2019off} and BCQ~\cite{kumar2019stabilizing}, respectively. Constrained Penalized Q-Learning~(\textbf{CPQ}~\cite{xu2022constraints}) aims to learn safe policy by penalizing the cost from the offline dataset. 
All Offline Safe RL baselines set an episodic cost constraint threshold $\kappa_c=1$. Based on the design of the safety cost introduced in Section~\ref{sec:env_setup}, when the episodic cost is lower than 1, it means that no critical violence, including collision and out-of-lane, occurred in this episode.


\subsection{Results and Analysis}
We design experiments and corresponding ablation studies to answer the following important research questions: 
\begin{itemize}
    \item \textbf{(RQ1)} How does FUSION perform with non-perfect offline data with diverse behavior policies from IDM and humans, compared with Safe Offline IL and RL baselines? 
    \item \textbf{(RQ2)} How does FUSION perform under unseen dynamics that the offline dataset does not cover, compared with all baselines? 
    \item \textbf{(RQ3)} Can FUSION consistently outperform other baselines and expert policies in diverse road contexts? 
    \item \textbf{(RQ4)} Do sequential modeling and causal representation learning benefit FUSION in capturing spatio-temporal dynamics contexts? 
\end{itemize}

For \textbf{RQ1} and \textbf{RQ2}, we compare FUSION with the baselines aforementioned in both policy mismatch and dynamics mismatch settings. 
The results in Table~\ref{tab:main_results} demonstrate the advantages of FUSION compared to baselines in both the safety cost and driving reward performance. (i) In the policy mismatch setting where the agent must overcome the suboptimality of the offline data, FUSION performs better in the reward~(driving efficiency), cost~(safety performance), and success rate. Notice that all the Safe IL baselines failed to learn a low-cost driving policy because these IL-based methods do not have explicit cost or reward feedback, and only fitting on those safe state and action transition pairs are insufficient to satisfy the safety requirements due to the imperfection of the offline demonstrations. Meanwhile, the Safe RL baselines seem to perform better, as they explicitly constrain the learned target policy with a preset cost threshold. The actor-critic framework that alternates between policy improvement and policy evaluation could implicitly guide the target policy to avoid some low-reward or high-cost behaviors. 
However, CPQ seems overly conservative in that it fails to balance efficiency and safety, thus always procrastinating near the starting zone to avoid getting a large cost penalty. 
On the other hand, ICIL, BNN, BEAR-Lag, and BCQ-Lag seem to have high success rates in policy mismatch settings, yet FUSION could still outperform them by a large margin~(over 10\%). 
(ii) In the dynamics mismatch case where the online testing environments have significantly different traffic dynamics and different types of roadblocks from training environments, the performance gap between our methods and other baselines even enlarged, for example, we can see that the success rate of Bear-Lag and BCQ-Lag drops by 40\%, and the evaluation cost of BCQ-Lag also violates the cost constraints. In contrast, although FUSION has a slightly lower reward than what it has in policy mismatch, the cost is still below the set threshold 1, and the success rate is also significantly higher than other baselines by more than 30\%.

\begin{figure}[h]
    \includegraphics[width=0.5\textwidth]{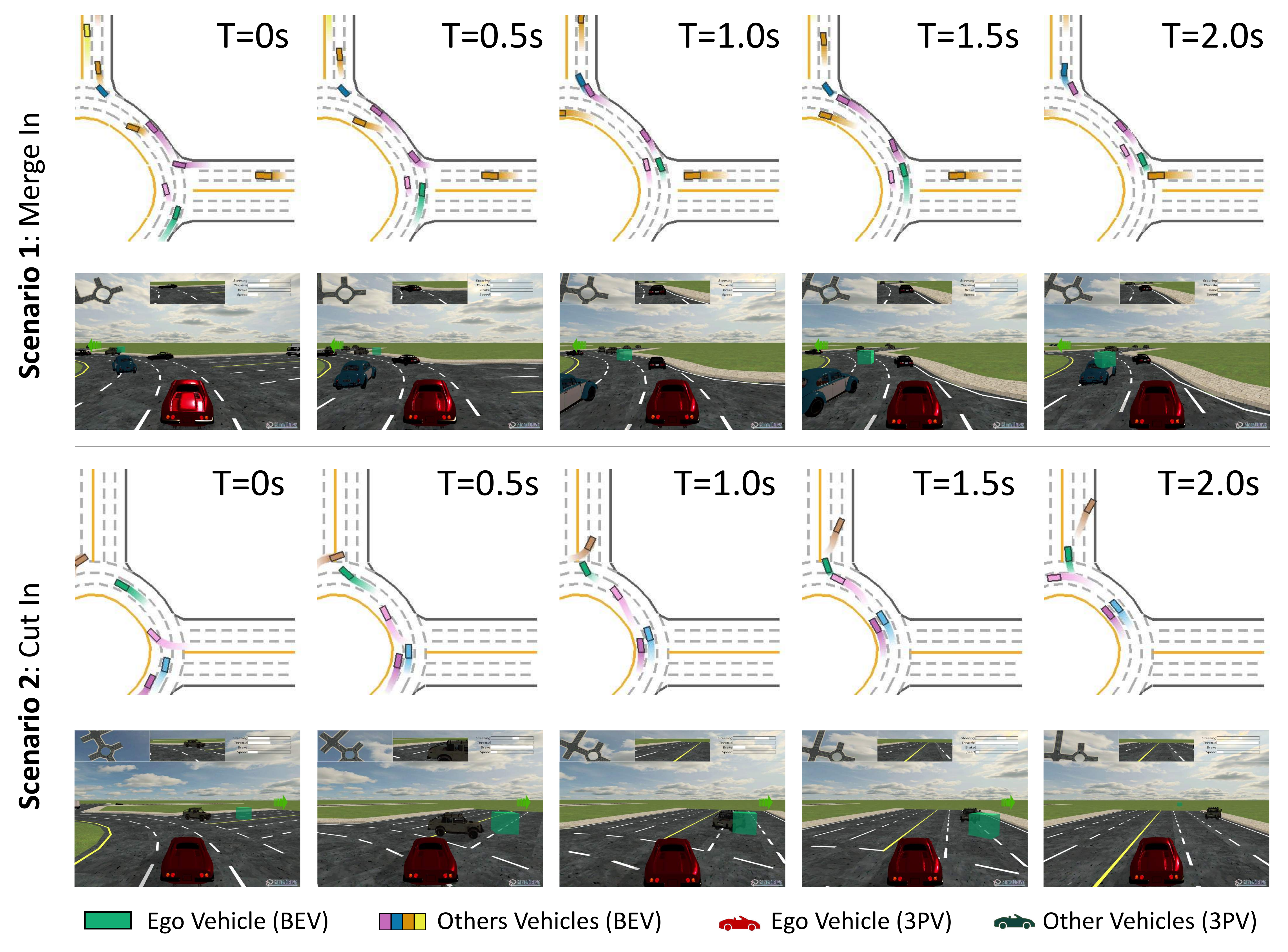}
    \vspace{-5mm}
    \caption{\revise{The figure shows both birds-eye-view (BEV) and third-person-view (3PV) images of two case studies in roundabouts. The first case is a merge-in behavior from normal traffic, and the ego vehicles controlled by FUSION will decelerate reasonably to keep the distance from the front vehicle. The second case is an adversarial driver trying to cut in from the wrong side of the roundabout exit, FUSION manages to yield to it safely. }}
    \label{fig:case-study}
    \vspace{-3mm}
\end{figure}

\begin{table*}[h]
    \centering
    \caption{Ablation studies on FUSION's variants to show the contribution of each module. \textbf{Bold} means the best.}
    \label{tab:ablation}
    \small{
    \begin{tabular}{c | c | c c c c | c} \toprule
        \textbf{Mismatch} & \textbf{Metrics} & \textbf{FUSION Short} & \textbf{FUSION w/o CEWM} & \textbf{FUSION w/o CBL} & \textbf{FUSION} & \textbf{Expert Policy} \\ \midrule
        \multirow{3}{*}{Policy} & Reward~($\uparrow$) & 100.86\scriptsize{$\pm$3.40} & 94.24\scriptsize{$\pm$6.50} & 104.54\scriptsize{$\pm$4.04} & \textbf{139.95\scriptsize{$\pm$4.24}} & 131.32\scriptsize{$\pm$29.60} \\
        & Cost~($\downarrow$) & 0.77\scriptsize{$\pm$0.09} & \textbf{0.67\scriptsize{$\pm$0.11}} & 3.46\scriptsize{$\pm$0.21} & \textbf{0.52\scriptsize{$\pm$0.06}} & 16.02\scriptsize{$\pm$9.46} \\
        & Succ. Rate~($\uparrow$) & 0.34\scriptsize{$\pm$0.07} & 0.41\scriptsize{$\pm$0.06} & 0.58\scriptsize{$\pm$0.09} & \textbf{0.90\scriptsize{$\pm$0.03}} & 0.81\scriptsize{$\pm$0.15} \\ \midrule
        \multirow{3}{*}{Dynamics}  & Reward~($\uparrow$) & 98.63\scriptsize{$\pm$2.36} & 81.70\scriptsize{$\pm$3.82} & 90.34\scriptsize{$\pm$4.28} & \textbf{117.40\scriptsize{$\pm$4.30}} & 129.71\scriptsize{$\pm$28.84} \\
        & Cost~($\downarrow$) & 0.79\scriptsize{$\pm$0.06} & \textbf{0.60\scriptsize{$\pm$0.04}} & 5.60\scriptsize{$\pm$0.32} & \textbf{0.90\scriptsize{$\pm$0.14}} & 17.58\scriptsize{$\pm$ 9.71} \\
        & Succ. Rate~($\uparrow$) & 0.34\scriptsize{$\pm$0.04} & 0.24\scriptsize{$\pm$0.04} & 0.08\scriptsize{$\pm$0.01} & \textbf{0.82\scriptsize{$\pm$0.04}} & 0.72\scriptsize{$\pm$0.20} \\ \bottomrule
    \end{tabular} 
    \vspace{-3mm}
    }
\end{table*}

\begin{figure}[h]
    \centering
    \includegraphics[width=0.5\textwidth]{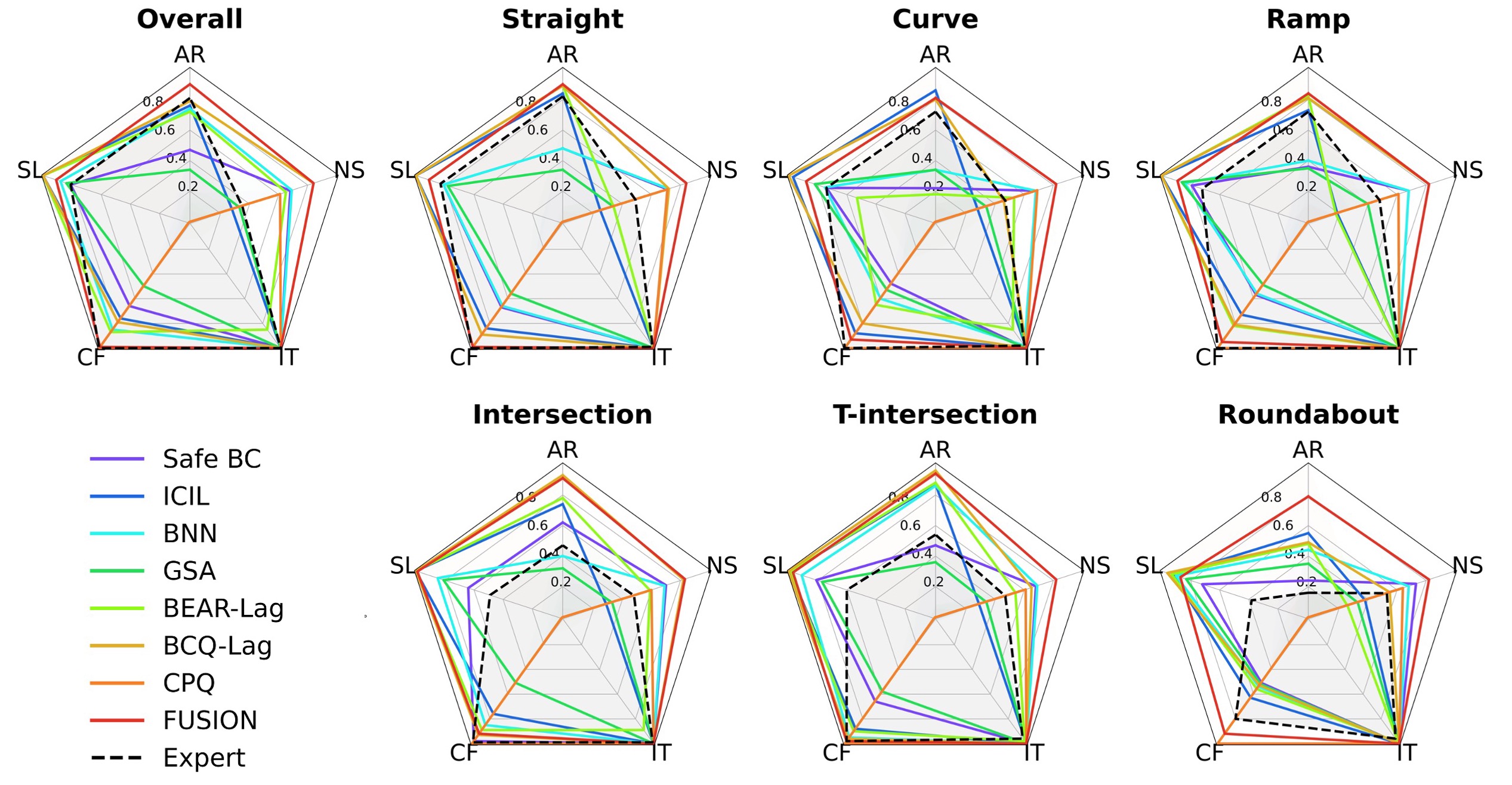}
    \vspace{-5mm}
    \caption{\revise{The figure shows the comparison of FUSION on different road configurations with baselines. The larger lidar plot on each coordinate stands for the safer performance in each safety metric. (\textbf{AR}: Arrival, \textbf{NS}: Not speeding, \textbf{IT}: In-time, \textbf{CF}: Collision-free, \textbf{SL}: Stay in-lane.)}}
    \label{fig:radar}
    \vspace{-3mm}
\end{figure}

\revise{For \textbf{RQ3}, we take a deeper look at the exact driving performance by case studies in Figure~\ref{fig:case-study}. 
We also provide comparisons of safety metrics in different road contexts as a radar plot in Figure~\ref{fig:radar}. 
The larger the pentagon is, the better overall safety performance it has. 
We calculate the safety metrics by the episode-wise frequency of five different safety behavior categories, }including 
(i) \textbf{AR}: arrival rate in all episodes; (ii) \textbf{NS}: not speeding in the episode, which counts the time step ratio in which the agent exceeds a speed limit of 40 \textit{kph} on the urban local roads; 
(iii) \textbf{IT}: in-time (complete the route within the time limit of 1,000 steps per episode);
(iv) \textbf{CF}: collision-free in a single episode; 
(v) \textbf{SL}: stay in-lane without violating the lane constraints. \revise{The result shows that our proposed FUSION agent can drive reasonably under complex contexts, especially in the hardest Roundabout environment. }

For \textbf{RQ4}, we provide additional ablation studies in Table~\ref{tab:ablation}. We compare FUSION with three of its variants: (i) \textbf{FUSION-Short}, which uses a shorter context in the safety-aware transformer to model the whole sequence; (ii) \textbf{FUSION w/o CEWM}, which does not consider the learning of the causal ensemble world model, and only uses the behavior cloning term as supervised signals; (iii) \textbf{FUSION w/o CBL}, which neglects safety-aware bisimulation learning.  The result confirms that the FUSION benefits from all its design, including the spatio-temporal information from CEWM and additional safety awareness in the transformer model via CBL. 

\begin{figure}
    \centering
    \includegraphics[width=0.48\textwidth]{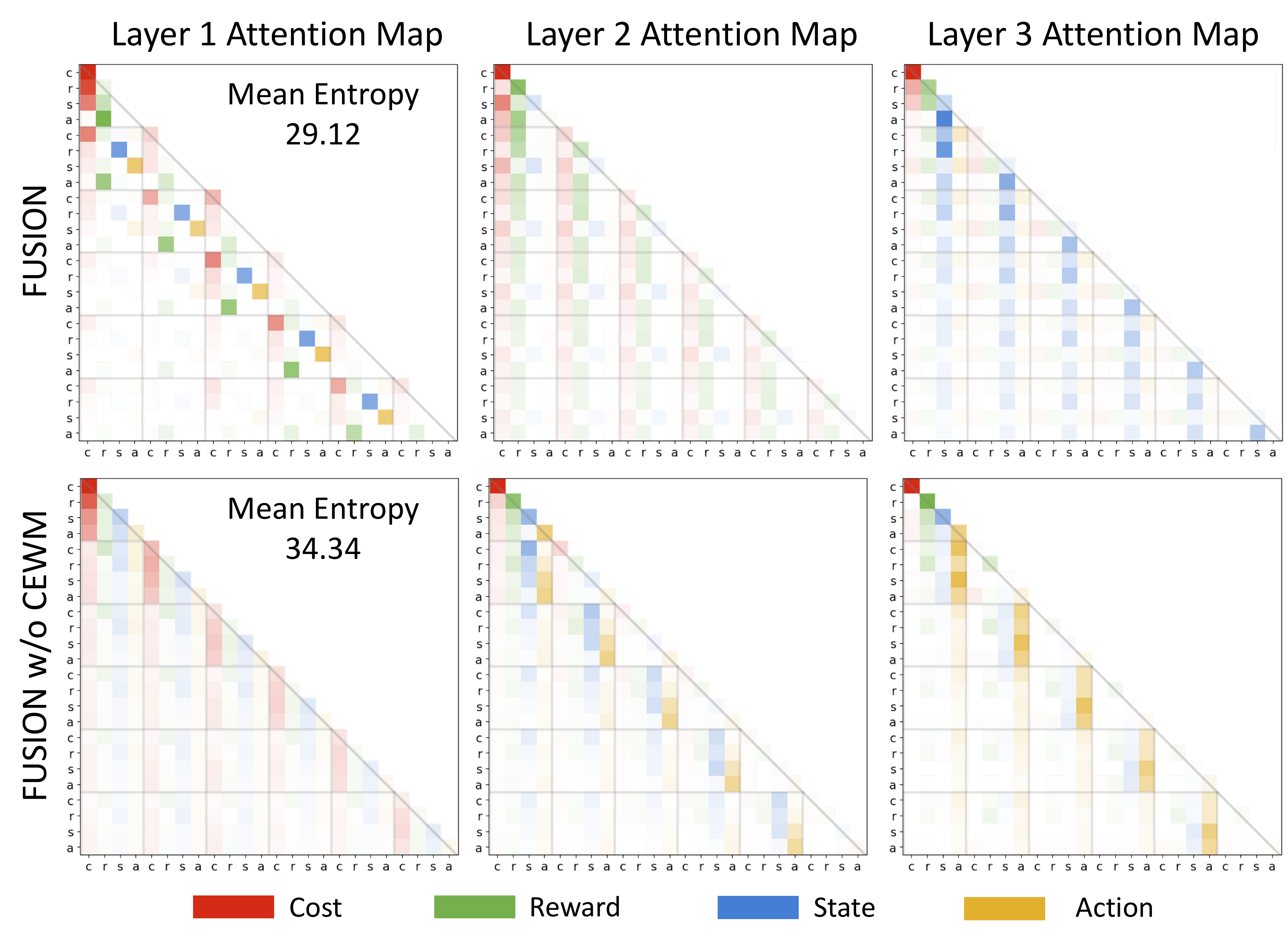}
    \vspace{-3mm}
    \caption{Visualization of average attention matrix over 30 trajectories. 
    We compare different layers of the attention map of two models: FUSION and FUSION w/o CEWM. 
    \revise{We compare the mean entropy through all three attention layers in one head of our Transformer encoder.
    The result shows that FUSION has a lower entropy than the ablation variant, which means its attention map is more sparse compared to the baselines without causal representation. }
    }
    \label{fig:attn}
    \vspace{-5mm}
\end{figure}

Furthermore, we visualize the \revise{normalized} attention map of FUSION's safety-aware causal transformer in Figure~\ref{fig:attn}. 
The x-axis represents the source (previous) nodes, and the y-axis represents the target (future) nodes. 
The attention map is a low-triangular matrix because only the tokens of previous timesteps affect the tokens in the future. 
We find that FUSION has a clear hierarchy in the attention map: (i) the attention map of the first layer is quite sparse, as FUSION only attends tokens from previous \textit{one} timestep, which essentially models the whole decision-making process in a Markovian manner. (ii) FUSION attends the preference tokens that include \textcolor{myred}{\textbf{cost-to-go}} and \textcolor{mygreen}{\textbf{reward-to-go}} to the future state and action tokens, trying to balance both for the decision-making process in a long horizon. (iii) FUSION captures world dynamics and policy by attending previous \textcolor{myblue}{\textbf{states}} to the future value prediction and action. 
Such semantically meaningful interpretation, as well as the heterogeneity of attention weights on different layers, indicate that FUSION benefits from CEWM by hierarchically capturing structural information reflected in the attention maps. 
On the contrary, as shown in the second row of Figure~\ref{fig:attn}, FUSION without CEWM \revise{has a higher average entropy among all the layers}, indicating that it does not capture the above sparsity and interpretability.
The reason is that the variant without CEWM \revise{ignores sequential awareness which has more informative training signals} during the offline training stage.

%% file: 5_conclusions.tex
\section{Conclusions}

In this paper, we propose FUSION, a trustworthy autonomous driving system with a causality-empowered safe reinforcement learning algorithm in an offline setting. 
We first design a safety-aware causal transformer termed CEWM to model the causal relationship between the state space, reward value, and cost value at different timesteps. 
\revise{Then we regularize the learned representation in CEWM with a CBL via safety-aware bisimulation in an implicit way, then greedily infer the action during online deployment. 
Exhaustive empirical results show that our method consistently outperforms several strong baselines of LfID and causal abstraction in diverse autonomous driving scenarios. 
We also conduct extensive case analysis to analyze the benefits of different modules that we design in FUSION and show a comprehensive and interpretable evaluation of FUSION.}
One potential limitation is that all the experiments are conducted in the portable MetaDrive simulator \revise{instead of more high-fidelity simulators like CARLA.}
\revise{Meanwhile, in the FUSION pipeline, CBL relies on a good estimation of transition dynamics, which in general requires good coverage and diversity in offline samples. }
In the future, it would be interesting to extend FUSION's framework to other autonomous vehicle platforms and tackle more challenging scenarios in multi-agent RL settings.